# Dynamic Modeling and Simulation of an Underactuated System

**Juan Libardo Duarte Madrid[1], P. A. Ospina-Henao[2] and E González Querubín[3]**

[1] Facultad de Ingeniería Mecatrónica, Universidad Santo Tomás, Bucaramanga, Colombia.
[2] Departamento de Ciencias Básicas, Universidad Santo Tomás, Bucaramanga, Colombia.
[3] Facultad de Ingeniería Mecatrónica, Universidad Santo Tomás, Bucaramanga, Colombia.

E-mail: juan.duarte03@ustabuca.edu.co, paolo.ospina@ustabuca.edu.co, edwin.gonzalez@ustabuca.edu.co

**Abstract**. In this paper, is used the Lagrangian classical mechanics for modeling the dynamics of an underactuated system, specifically a rotary inverted pendulum that will have two equations of motion. A basic design of the system is proposed in SOLIDWORKS 3D CAD software, which based on the material and dimensions of the model provides some physical variables necessary for modeling. In order to verify the results obtained, a comparison the CAD model simulated in the environment SimMechanics of MATLAB software with the mathematical model who was consisting of Euler-Lagrange's equations implemented in Simulink MATLAB, solved with the ODE23tb method, included in the MATLAB libraries for the solution of systems of equations of the type and order obtained. This article also has a topological analysis of pendulum trajectories through a phase space diagram, which allows the identification of stable and unstable regions of the system.

## 1. Introduction

Nowadays, the underactuated mechanical systems [1] are generating interest among researchers of modern control theories. This interest is that these systems have similar problems to those found in industrial applications, such as disturbances, instabilities and nonlinear behavior in some conditions of operation. The rotary inverted pendulum [2] is a clear example of an underactuated mechanical system; this is a mechanism of two degrees of freedom (DOF) and two rotational joints. It consists of three main elements: a motor, a rotational arm and a pendulum [3]. The motor shaft is connected to one end of the rotational arm making this fully rotate on a horizontal plane; the other end of the arm has connected the pendulum that freely rotates 360 degrees in a vertical plane. Despite being a purely academic level, this system is helpful to study, apply and analyze different modeling strategies.

Some industrial applications [4] that present the disadvantages mentioned above, are in fields such as robotics, robots balance, biped robots [5], robotic arms; telecommunications, satellite positioning; transportation, Segway [6], [7], stability of ships and submarines, iBot [8], Self-balancing unicycle; and field monitoring, drones.

For the accomplishment of this work, initially there was realized a bibliographic review of similar projects [9], [10], [11], in order to determine the dimensions and suitable material for the later design of the prototype in SOLIDWORKS 3D Computer Aided Engineering (CAD) software. Then, was modeled the dynamics of the system by Lagrangian mechanics, obtaining two equations of motion; one corresponds to the arm and other one to the pendulum. To validate the equations obtained, multiple simulations were made in the MATLAB software in order to observe their behavior. Thus, it was possible to verify in a graphic mode, the waveform of the angular position and velocity of the arm and the pendulum.

A fundamental part in the stage of simulations was to import the CAD model made in SOLIDWORKS to environment of SimMechanics [12] in MATLAB. Immediately afterwards, the equations obtained were solved by the ODE23tb (Ordinary Differential Equations) method belonging to MATLAB. The use of block diagram in the environment Simulink of MATLAB allowed a representation of the equations of motion. All this had the purpose of elaborating a comparison of the response of both the CAD model and the mathematical model obtained by the Lagrange formalism.

Finally, through the effective potential of the system the phase space diagram [13] was plotted, in this one can see the trajectories of the pendulum and to study the critical points that correspond to the maximum and minimum of that potential.

In Section II is defined in a broader way the rotational inverted pendulum, in addition a table is shown that contains some variables that will be used throughout the article along with a basic design of the system. In section III It provides a brief definition of an underactuated systems. Next, in section IV presents the advantage of making use of Euler-Lagrange formalism. Continuing, in Section V it shows the dynamic modeling of the proposed rotary inverted pendulum. In section VI activities are defined for performing proposed simulations as a method of validating results. In section VII the results obtained are shown with a respective analysis of them. Finally, section VIII contains the relevant conclusions of the article.

**2. Prototype Proposed**

The pendulum is a physical object consisting of a mass joined by means of a bar to a pivot. The system is free to move, realizing a balancing about its point of stable equilibrium; namely, in its hanging position, in which it will be as long as there are no forces that influence its nature. In 1991 [14], there was created the system that shows itself in the Figure 1, this device is known as Furuta's pendulum, referring to your creator Katsuhisa Furuta. This system consists of two links only, the first one of them is known as arm, which is an actuated element. Since this one turns freely in a horizontal plane with the help of an electrical engine of direct current, the second one knows himself as pendulum; this one is an underactuated element that turns freely in a vertical plane for action of the movement transmitted by the arm. Therefore, it is a question of an underactuated system with two GDL and one control input [15].

The rotary inverted pendulum or Furuta's pendulum is of great help in the academy for the analysis and study of strategies of classic, modern, not linear and advanced control; also, it is used with didactic intentions for the education of the theory of control, classic mechanics and system identification.

This section presents the basic design created in SOLIDWORKS and some variables needed for modeling.

**Table 1.** Variable System

| Physical characteristic | Symbology | Units |
|---|---|---|
| Acceleration of gravity | $g$ | $m/s^2$ |
| Angular position of the arm | $\theta_0$ | $rad$ |
| Angular position of the pendulum | $\theta_1$ | $rad$ |
| Arm length | $L_0$ | $m$ |
| Generalized coordinates | $q$ | $rad$ |
| Generalized velocities | $\dot{q}$ | $rad/s$ |
| Kinetic Energy | $K$ | $J$ |
| Lagrangian | $\mathcal{L}$ | $J$ |
| Location of the center of mass of the pendulum | $l_1$ | $m$ |
| Moment of inertia of the arm | $I_0$ | $kg.m^2$ |
| Moment of inertia of the pendulum | $I_1$ | $kg.m^2$ |
| Pendulum length | $L_1$ | $m$ |
| Pendulum mass | $m_1$ | $kg$ |
| Potential Energy | $U$ | $J$ |

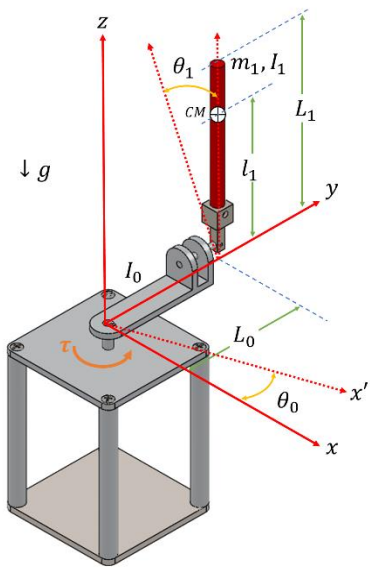

**Figure 1.** Rotary inverted pendulum

### 3. Underactuated Systems

In the study of mechanisms are acquired two concepts very important such as the direct and indirect action. The first consists of movement of elements by action of an actuator, while the second consists of the action of motion transmitted by another interconnected element. Such movements are known as DOF, so that mechanical systems or mechanisms can be classified depending on the number of DOF and the number of actuators. The fully actuated mechanical systems are those having the same number of DOF and actuators. Underactuated mechanical systems are those with fewer actuators than DOF [16]. It is important to highlight the advantages of underactuated systems, since if they do not have advantages over fully actuated mechanical systems, it will not make sense its development. The main advantages present in underactuated systems are energy saving and control efforts. However, these systems are intended to perform the same functions of fully actuated systems without their disadvantages.

### 4. Euler-Lagrange

The dynamic equations of any mechanical system can be obtained from the known Newtonian classical mechanics, the drawback of this formalism is the use of the variables in vector form, complicating considerably the analysis when increasing the joints or there are rotations present in the system. In these cases, it is favorable to employ the Lagrange equations, which have formalism of scale, facilitating the analysis for any mechanical system.

In order to use Lagrange equations, it is necessary to follow four steps:

- Calculation of kinetic energy.
- Calculation of the potential energy.
- Calculation of the Lagrangian.
- Solve the equations.

Where the kinetic energy can be both rotationally and translational, this form of energy may be a function of both the position and the speed $K(q(t), \dot{q}(t))$.

The potential energy is due to conservative forces as the forces exerted by springs and gravity, this energy is in terms of the position $U(q(t))$.

Is defined the Lagrangian as

$$\mathcal{L} = K - U \tag{1}$$

Therefore, the Lagrangian in general terms is defined of the following way

$$\mathcal{L}(q(t), \dot{q}(t)) = K(q(t), \dot{q}(t)) - U(q(t)) \tag{2}$$

Finally, are defined the Euler-Lagrange equations for a system of $n$ DOF as follows

$$\frac{d}{dt}\left(\frac{\partial \mathcal{L}(q, \dot{q})}{\partial \dot{q}_i}\right) - \frac{\partial \mathcal{L}(q, \dot{q})}{\partial q_i} = \tau_i \tag{3}$$

Where $i = 1, \ldots, n$, $\tau_i$ are the forces or pairs exerted externally in each joint, besides nonconservative forces such as friction, resistance to movement of an object within a fluid and generally those that depend on time or speed. It will be obtained an equal number of dynamic equations and DOF.

**5. Modeling of Rotary Inverted Pendulum**

Was performed the dynamic modeling of the model shown in the Figure 1. It is necessary to make an energy analysis. Therefore, initially the kinetic energy of each link is analyzed, so it can be identified which kinetic energies (rotational and translational) were present in each link.

*5.1. Kinetic Energy*

The kinetic energy of the system consist of a translational and rotational component for the pendulum and rotational component for the arm

$$K = \frac{1}{2}mv^2 + \frac{1}{2}I\omega^2 \tag{4}$$

Where $m$ is the mass of the body, $v$ the linear velocity, $I$ the moment of inertia, $\omega$ the angular velocity and $K$ the kinetic energy. In this case, there are two bodies, the arm and the pendulum. The kinetic energy of the arm is

$$K_0 = \frac{1}{2}I_0\dot{\theta}_0^{\,2} \tag{5}$$

The kinetic energy of the pendulum is

$$K_1 = \frac{1}{2}m_1v_1^{\,2} + \frac{1}{2}I_1\dot{\theta}_1^{\,2} \tag{6}$$

The total energy of the system is

$$K_T = \frac{1}{2}I_0\dot{\theta}_0^{\,2} + \frac{1}{2}m_1v_1^{\,2} + \frac{1}{2}I_1\dot{\theta}_1^{\,2} \qquad (7)$$

*5.2. Potential Energy*
This system only store gravitational potential energy in the pendulum

$$U = mgh \qquad (8)$$

The arm has in its nature a rotational movement in a horizontal plane, therefore it does not have height change in its center of mass, providing equal 0 component in equation (8) resulting in a potential energy zero .
The potential energy of the pendulum is

$$U_1 = m_1gl_1(\cos\theta_1 - 1) \qquad (9)$$

Where $g$ represents the value of gravity. The total potential energy of the system is (9)

$$U_T = m_1gl_1(\cos\theta_1 - 1) \qquad (10)$$

*5.3. Position of the pendulum*
Because the pendulum is a rigid body, the required position is the its center of mass

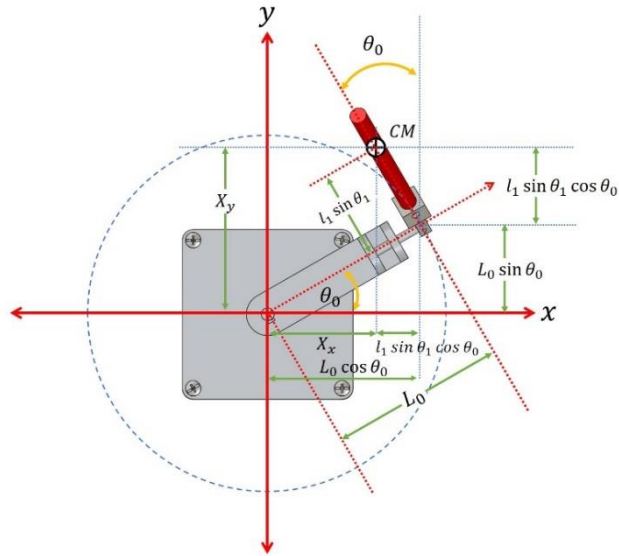

**Figure 2.** Projection arm and pendulum in the $xy$ plane

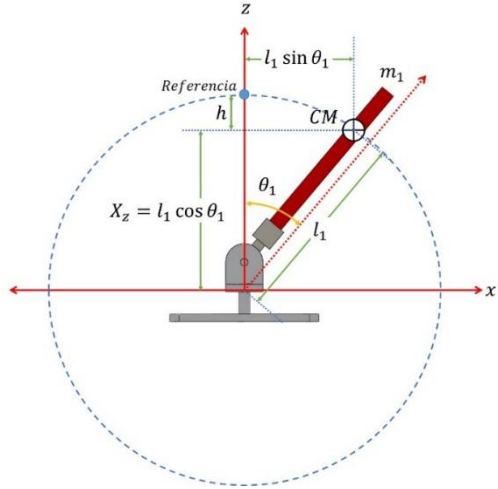

**Figure 3.** Projection pendulum in the $xz$ plane

Then, it shows components the position of the center of mass

$$X_x = L_0 \cos \theta_0 - l_1 \sin \theta_1 \sin \theta_0 \tag{11}$$

$$X_y = L_0 \sin \theta_0 + l_1 \sin \theta_1 \cos \theta_0 \tag{12}$$

$$X_z = l_1 \cos \theta_1 \tag{13}$$

The position of the center of mass

$$X_{CM} = [X_x \quad X_y \quad X_z]^T \tag{14}$$

*5.4. Linear Speed of the center of mass*
The speed is defined as the derivative of the position with respect to time

$$\frac{d}{dt}(X_{CM}) = V_{CM} \tag{15}$$

$$\dot{X}_{CM} = V_{CM} = [\dot{X}_x \quad \dot{X}_y \quad \dot{X}_z]^T \tag{16}$$

The velocity components are obtained by differentiating each component position (11), (12) and (13) respectively

$$\dot{X}_x = -\dot{\theta}_0 L_0 \sin \theta_0 - l_1(\dot{\theta}_0 \sin \theta_1 \cos \theta_0 + \dot{\theta}_1 \cos \theta_1 \sin \theta_0) \tag{17}$$

$$\dot{X}_y = \dot{\theta}_0 L_0 \cos \theta_0 + l_1(\dot{\theta}_1 \cos \theta_1 \cos \theta_0 - \dot{\theta}_0 \sin \theta_1 \sin \theta_0) \tag{18}$$

$$\dot{X}_z = -\dot{\theta}_1 l_1 \sin \theta_1 \tag{19}$$

The linear velocity of the center of mass can be expressed as follows vector form

$$V_{CM}{}^2 = \begin{bmatrix} \dot{X}_x & \dot{X}_y & \dot{X}_z \end{bmatrix} \begin{bmatrix} \dot{X}_x \\ \dot{X}_y \\ \dot{X}_z \end{bmatrix} = \dot{X}_x{}^2 + \dot{X}_y{}^2 + \dot{X}_z{}^2 \qquad (20)$$

They are calculated separately each of the components to the square of the velocity of the center of mass

$$\dot{X}_x{}^2 = \dot{\theta}_0{}^2 L_0{}^2 \sin^2 \theta_0 - 2(\dot{\theta}_0 L_0 \sin \theta_0)\left(l_1(\dot{\theta}_0 \sin \theta_1 \cos \theta_0 + \dot{\theta}_1 \cos \theta_1 \sin \theta_0)\right) \qquad (21)$$
$$+ l_1{}^2(\dot{\theta}_0 \sin \theta_1 \cos \theta_0 + \dot{\theta}_1 \cos \theta_1 \sin \theta_0)^2$$

$$\dot{X}_y{}^2 = \dot{\theta}_0{}^2 L_0{}^2 \cos^2 \theta_0 + 2(\dot{\theta}_0 L_0 \cos \theta_0)\left(l_1(\dot{\theta}_1 \cos \theta_1 \cos \theta_0 - \dot{\theta}_0 \sin \theta_1 \sin \theta_0)\right) \qquad (22)$$
$$+ l_1{}^2(\dot{\theta}_1 \cos \theta_1 \cos \theta_0 - \dot{\theta}_0 \sin \theta_1 \sin \theta_0)^2$$

$$\dot{X}_z{}^2 = \dot{\theta}_1{}^2 l_1{}^2 \sin^2 \theta_1 \qquad (23)$$

Finally, the following expression is obtained for the linear velocity

$$V_{CM}{}^2 = L_0{}^2 \dot{\theta}_0{}^2 + l_1{}^2(\dot{\theta}_1{}^2 + \dot{\theta}_0{}^2 \sin^2 \theta_1) + 2 L_0 l_1 m_1 \dot{\theta}_0 \dot{\theta}_1 \cos \theta_1 \qquad (24)$$

*5.5. Energy System*

The total energy of the system is

$$K_T = K_0 + K_1 \qquad (25)$$

Substituting (5) and (6) in (25) is obtained

$$K_T = \frac{1}{2} I_0 \dot{\theta}_0{}^2 + \frac{1}{2} I_1 \dot{\theta}_1{}^2 + \frac{1}{2} m_1 V_{CM}{}^2 \qquad (26)$$

Substituting (24) in (26) is obtained the total kinetic energy

$$K_T = \frac{1}{2} I_0 \dot{\theta}_0{}^2 + \frac{1}{2} I_1 \dot{\theta}_1{}^2 + \frac{1}{2} m_1 \{ L_0{}^2 \dot{\theta}_0{}^2 + l_1{}^2(\dot{\theta}_1{}^2 + \dot{\theta}_0{}^2 \sin^2 \theta_1) + 2 L_0 l_1 m_1 \dot{\theta}_0 \dot{\theta}_1 \cos \theta_1 \} \qquad (27)$$

Is shown in (10) the potential energy of the system

*5.6. Euler-Lagrange Equations*

The Lagrangian of the system is

$$\mathcal{L} = K_T - U_T \qquad (28)$$

Substituting (10) and (27) in (28) the Lagrangian of the system is obtained

$$\mathcal{L} = \frac{1}{2} I_0 \dot{\theta}_0{}^2 + \frac{1}{2} I_1 \dot{\theta}_1{}^2 + \frac{1}{2}\left(L_0{}^2 m_1 \dot{\theta}_0{}^2\right) + \frac{1}{2}\left(l_1{}^2 m_1 \dot{\theta}_1{}^2\right) + \frac{1}{2}\left(l_1{}^2 m_1 \dot{\theta}_0{}^2 \sin^2 \theta_1\right) \qquad (29)$$
$$+ L_0 l_1 m_1 \dot{\theta}_0 \dot{\theta}_1 \cos \theta_1 + m_1 g l_1 (1 - \cos \theta_1)$$

As there are two DOF, the Euler-Lagrange equations have the following form

$$\frac{d}{dt}\left(\frac{\partial \mathcal{L}}{\partial \dot{\theta}_0}\right) - \frac{\partial \mathcal{L}}{\partial \theta_0} = \tau \tag{30}$$

$$\frac{d}{dt}\left(\frac{\partial \mathcal{L}}{\partial \dot{\theta}_1}\right) - \frac{\partial \mathcal{L}}{\partial \theta_1} = 0 \tag{31}$$

Where $\tau$ is the torque of motor, solving (30) and (31) we obtain the equations of motion which are given by

$$I_0\ddot{\theta}_0 + L_0^2 m_1 \ddot{\theta}_0 + \{l_1^2 m_1 (\ddot{\theta}_0 \sin^2\theta_1 + 2\dot{\theta}_0\dot{\theta}_1 \sin\theta_1 \cos\theta_1)\} + \{L_0 l_1 m_1 (\ddot{\theta}_1 \cos\theta_1 - \dot{\theta}_1^2 \sin\theta_1)\} = \tau \tag{32}$$

$$I_1\ddot{\theta}_1 + l_1^2 m_1 \ddot{\theta}_1 + L_0 l_1 m_1 \ddot{\theta}_0 \cos\theta_1 - l_1^2 m_1 \dot{\theta}_0^2 \sin\theta_1 \cos\theta_1 - m_1 g l_1 \sin\theta_1 = 0 \tag{33}$$

Where (32) is the equation of motion of the arm and (33) of the pendulum.

**6. Simulation**
Although the Euler-Lagrange formalism ensures a high degree of approximation of the mathematical models, is essential do comparisons to validate these results. Are followed the next steps for verification of modeling:

   *1)* Represent the system equations in the state space.
   *2)* Define an experiment with initial conditions, natural interactions and external forces.
   *3)* Import the CAD model of SOLIDWORKS in SimMechanics-MATLAB.
   *4)* Add the necessary blocks to obtain the desired graphic model and applying external forces.
   *5)* Simulating experiment.
   *6)* Export the results of SimMechanics to Workspace MATLAB.
   *7)* Implement a block diagram Simulink-MATLAB to solve the equations.
   *8)* Export the solutions to the equations to Workspace MATLAB.
   *9)* Graphing and overlay solutions.

As can be seen in the steps above, the simulation of the model was divided in two stages: first, simulate the CAD model initially designed and the second implement the equations obtained.

**7. Results**
Then, it presents each of the steps mentioned in the previous section.
   *1)* Starting from [17]

$$M(q)\ddot{q} + C(q,\dot{q})\dot{q} + G(q) = \tau \tag{34}$$

Therefore, the equation (34) is the dynamic equation for mechanical systems of n DOF. Where, $M$ is the matrix of inertia, $C$ is the centrifugal and Coriolis matrix, $G$ the vector of gravity and $\tau$ external forces.
Taking the equations of motion (32) and (33) and replacing in (34) is obtained a matrix representation

$$\begin{bmatrix} I_o + m_1 L_0^2 + l_1^2 m_1 \sin^2\theta_1 & L_0 l_1 m_1 \cos\theta_1 \\ L_0 l_1 m_1 \cos\theta_1 & I_1 + m_1 l_1^2 \end{bmatrix} \begin{bmatrix} \ddot{\theta}_0 \\ \ddot{\theta}_1 \end{bmatrix} + \begin{bmatrix} 2l_1^2 m_1 \sin\theta_1 \cos\theta_1 \dot{\theta}_1 & -L_0 l_1 m_1 \sin\theta_1 \dot{\theta}_1 \\ -l_1^2 m_1 \sin\theta_1 \cos\theta_1 \dot{\theta}_0 & 0 \end{bmatrix} \begin{bmatrix} \dot{\theta}_0 \\ \dot{\theta}_1 \end{bmatrix} + \begin{bmatrix} 0 \\ -gl_1 m_1 \sin\theta_1 \end{bmatrix} = \begin{bmatrix} \tau \\ 0 \end{bmatrix} \tag{35}$$

The matrix $M(q)$ is important for the dynamic modeling and for the design of controllers. This matrix has a great relationship with the kinetic energy, also the inertia matrix is a symmetric, positive and square matrix of $n \times n$, whose elements depend only on the generalized coordinates.

Centrifugal and Coriolis matrix $C(q,\dot{q})$ it is important in the study of stability in control systems, mechanical systems, among others. This matrix is square of $n \times n$ and has dependence in its elements of the generalized coordinates and velocities.

The gravity vector $G(q)$ it is present in mechanical systems without counterweights or springs, in turn is in systems with displacement off the horizontal plane. This vector is of $n \times 1$ and has only reliance on joint positions.

*2)* It defined that the system would have the initial conditions shown in Figure 5, in addition to being subject to effects of gravity and to a torque step 0.2 seconds in the end of the arm that connects to the motor shaft. Finally, a simulation interval 5 seconds was established.

*3)* Was imported the CAD model in SimMechanics with the following code line: *mech_import('CADModel_Pendulum.xml')*. The initial position of the system is Figure 5 a).

*4)* It was necessary add a few block the diagram SimMechanics, since the CAD model is only under the effect of gravity and not have some kind of movement, it blocks provide the step of torque arm included to start rotating, besides adding blocks to the sensing of angular displacement and velocities in an interval of 5 second of test.

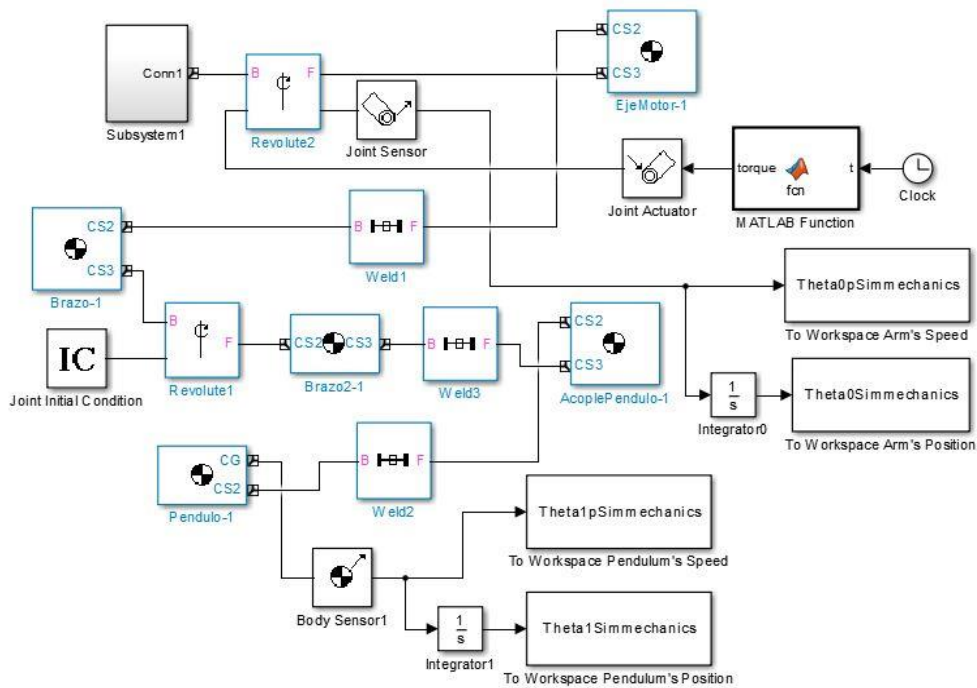

**Figure 4.** Final block diagram in SimMechanics

The Figure 4 shows the block diagram implemented for the simulation of the CAD model. The block named Subsystem1 contains the majority of the blocks generated of the CAD model by SimMechanics, such blocks are not necessary for the simulation; others blocks generated by SimMechanics are Revolute2, EjeMotor-1, Weld1, Brazo-1, Revolute1, Brazo2-1, Weld3, AcoplePendulo-1, Weld2 and Pendulo-1. The most important elements for the simulation are the blocks Revolute; Revolute2 is the joint of the arm, theoretically it is there where there would be connected the electrical engine, which in this case is simulated by a Joint Actuator configured as a force generator. The block MATLAB Function contains a small code that generates a torque of 0.5 Nm in an interval of 0.2 seconds; to simulate the motor encoder the Joint Sensor block is connected, of which there is obtained the speed of arm it $\dot{\theta}_0$

(Figure 11), and with an of integration block (Integrator0) the position of same one it $\theta_0$ (Figure 9). For the case of the Revolute1 there is a bit different the way of obtaining the variables of interest, this difference owes to the CAD design in SOLIDWORKS, since both the arm and the pendulum have design and couplings different. In Revolute1 there connects a block called Joint Initial Condition, in which it is possible to change the initial position of the pendulum, in this link we will not have any engine, since as mentioned in the section II it is a question of the underactuated element. On the other hand, it is necessary to simulate an encoder to know the position of the pendulum in any instant of time, for this implemented the block Body Sensor1, which is used to measure the position of the center of mass of the pendulum. So that, there is obtained initially the angular speed of the pendulum it $\dot{\theta}_1$ (Figure 10) and of equal way that with the arm, by means of an integrator (Integrator1) obtains the angular position it $\theta_1$ (Figure 8).

  5) Some moments of the position of the system were taken in the time interval defined.

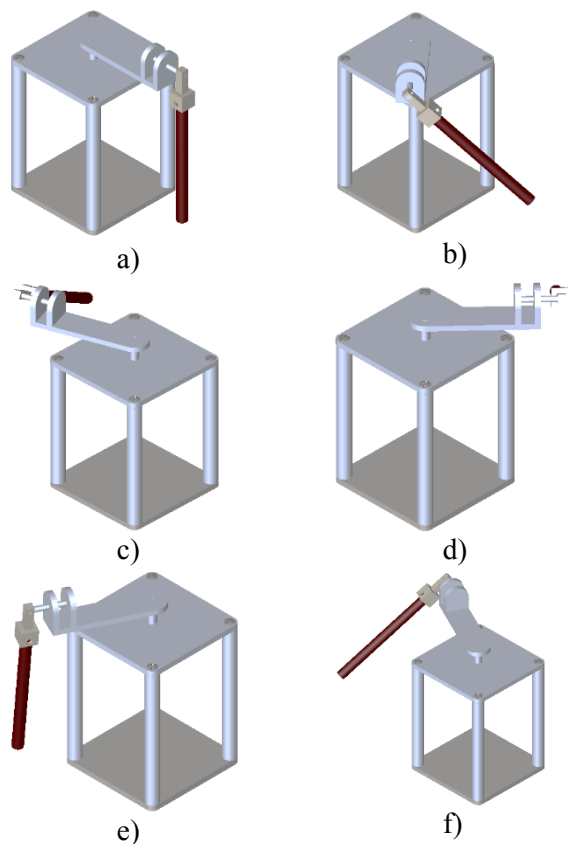

**Figure 5.** Experiment in SimMechanics
a) t = 0 s, b) t = 0.212227 s, c) t = 0.798791 s, d) t = 1.22854 s, e) t = 3.33321 s, f) t = 5 s

  6) To export system solutions to Workspace, is need to add the blocks with the name To Workspace located in Simulink library, which will create a cell for each solution, with the respective data and time in which such data is obtained.
  7) The implementation of the block diagram was performed to numerically solve the equations of the system, thus obtaining values $\theta_0, \theta_1, \dot{\theta}_0, \dot{\theta}_1$ along the defined time interval. The diagram implemented is the follows:

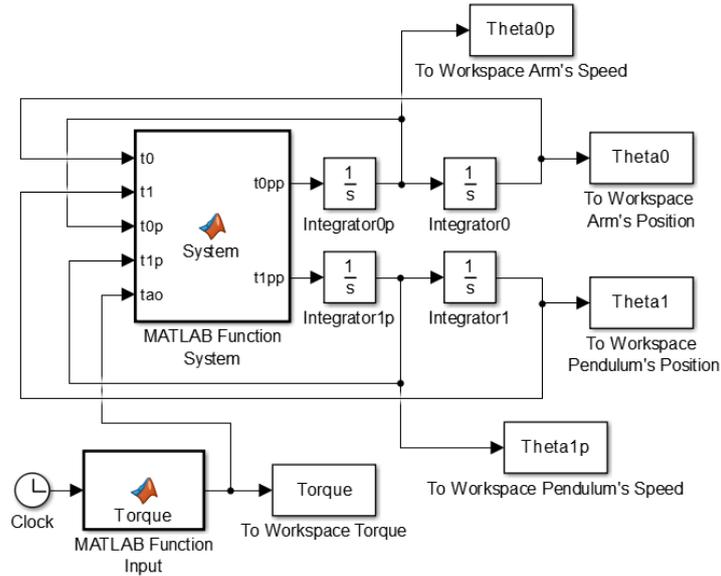

**Figure 6.** Block diagram of Euler-Lagrange's equations

Wherein, the block MATLAB Function Input, has the input applied to the system, which in this case is a torque step of 0.2 seconds with amplitude of 0.5 Nm, this being the maximum torque generated by the motor. The block called MATLAB Function System, contains the mathematical model in its matrix representation, with five inputs ($\theta_0, \theta_1, \dot{\theta}_0, \dot{\theta}_1, \tau$) and two outputs ($\ddot{\theta}_0, \ddot{\theta}_1$). The integrators are used to obtain speed and position respectively. Internally, the block uses the ODE23tb method for solving equations.

Numerical values necessary to solve the Euler-Lagrange's equations of motion are as follows

**Table 2.** Numerical Variable System

| Physical characteristic | Values |
|---|---|
| Acceleration of gravity | $9.81 \ m/s^2$ |
| Arm length | $0.201 \ m$ |
| Location of the center of mass of the pendulum | $0.154985 \ m$ |
| Moment of inertia of the arm | $0.0052 \ kg.m^2$ |
| Moment of inertia of the pendulum | $0.0023 \ kg.m^2$ |
| Pendulum length | $0.30997 \ m$ |
| Pendulum mass | $0.2866 \ kg$ |

Was made an approximation in the calculation of the moments of inertia, both links are taken as constant circular bars and invariant mass

$$I_0 = \frac{1}{3} m_0 L_0^2 \tag{36}$$

$$I_1 = \frac{1}{12} m_1 L_1^2 \tag{37}$$

Where (36) is the moment of inertia of the arm; measured from the end connected to the motor shaft to the opposite end. While (37) is the pendulum moment of inertia; measured pendulum from the center of mass.

*8)* The way to export the data to Workspace of block diagram above is performed with the same aggregate block in step *6)*.

*9)* Taking the exported data in points *6)* and *8)* the following graphs were made.

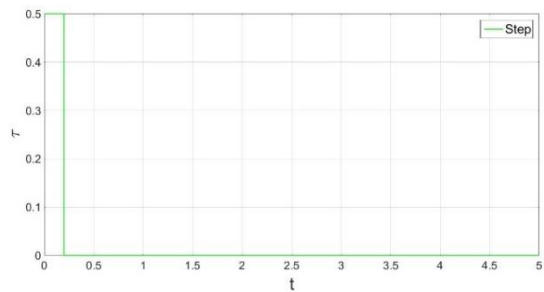

**Figure 7.** Step torque

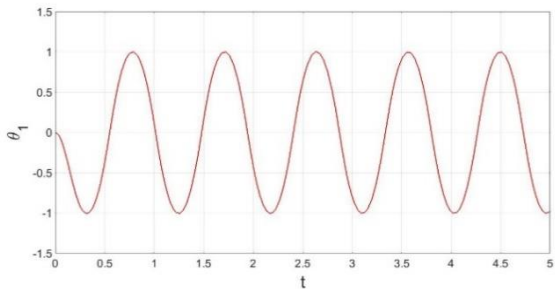

**Figure 8.** Angular position of pendulum with CAD model

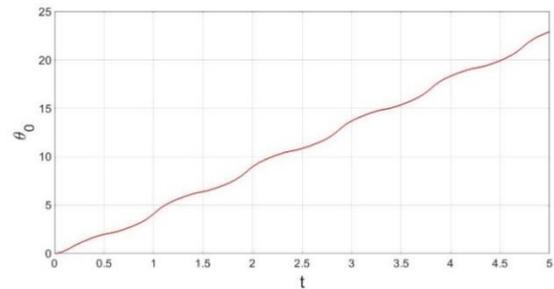

**Figure 9.** Angular position of arm with CAD model

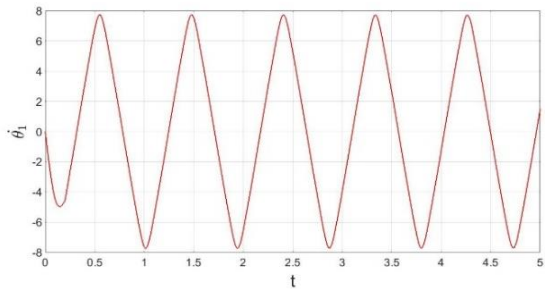

**Figure 10.** Angular velocity of pendulum with CAD model

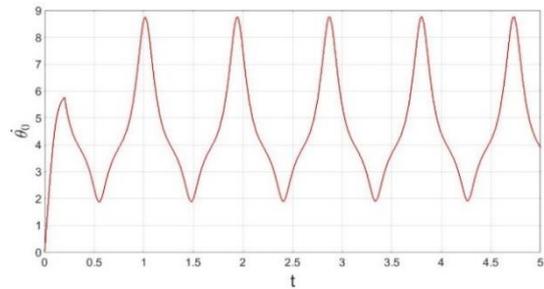

**Figure 11.** Angular velocity of arm with CAD model

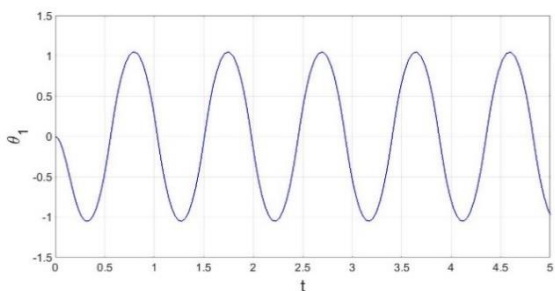

**Figure 12.** Angular position of pendulum with Euler-Lagrange's equations

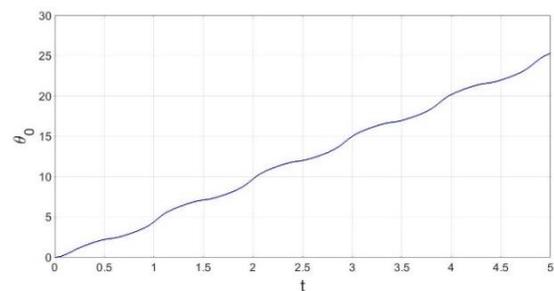

**Figure 13.** Angular position of arm with Euler-Lagrange's equations

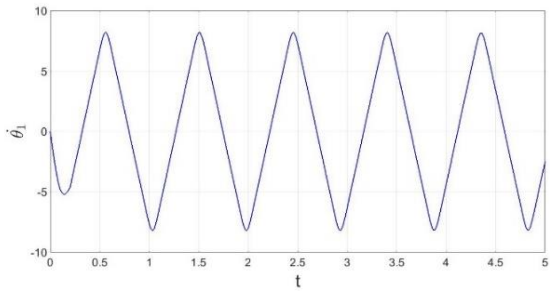
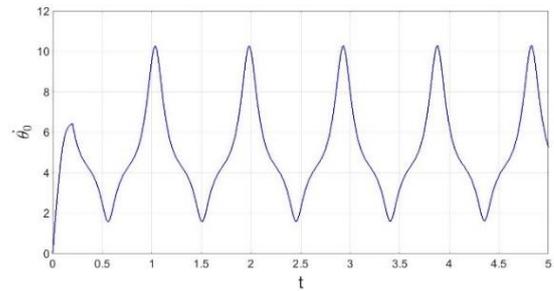

**Figure 14.** Angular velocity of pendulum with Euler-Lagrange's equations

**Figure 15.** Angular velocity of arm with Euler-Lagrange's equations

It is noted that the system behavior is the same in both simulations. A superposition of results was performed by way of display the error between the CAD model simulated and the Euler-Lagrange equation, with the following results.

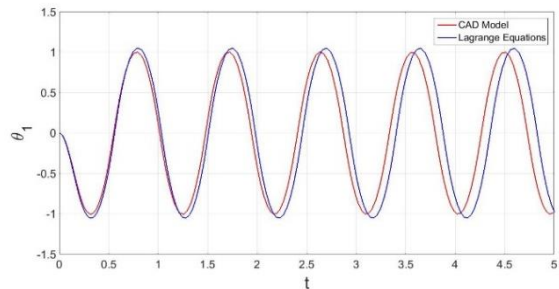

**Figure 16.** Comparison results angular position of pendulum

Figure 16 allows viewing behavior overlapping displacement of the center of mass in the pendulum. The red line represents the CAD model while the blue is the result of $\theta_1$ graph the solution over time.

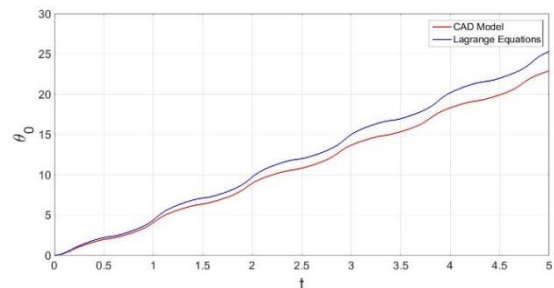

**Figure 17.** Comparison results angular position of arm

Figure 17 is overlaying the results to simulate and plot $\theta_0$ function of time, where the red line is the result of simulating in SimMechanics the CAD model and the blue line is the solution of the Euler-Lagrange's equation.

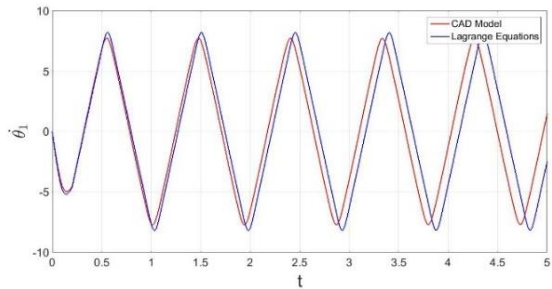

**Figure 18.** Comparison results angular velocity of pendulum

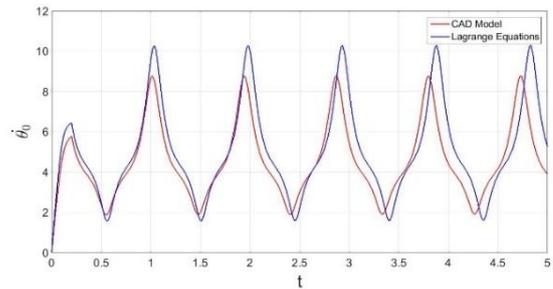

**Figure 19.** Comparison results angular velocity of arm

Similarly as with Figure 16 and 17, the Figure 18 and 19 illustrate the behavior of the angular velocity with respect to time for the pendulum and the arm respectively.

It is important to note that the graphs do not have a zero error due to the approximations that were made to the calculation of the moments of inertia of each link; the difference is that the SOLIDWORKS software performs a more accurate calculation of the moments of inertia, because the software considers the geometry of each link.

Then, the pendulum trajectory analysis shown, which part of the potential energy in the center of mass as shown in (9).

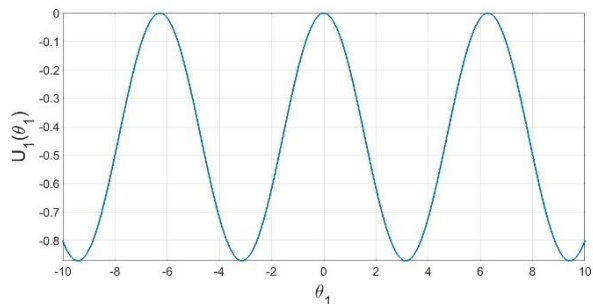

**Figure 20.** Potential energy

The phase space diagram is obtained when all the trajectories of the pendulum are gathered, which shows which are the critical points of the system. In this case, it's know that the pendulum has two critical points, a stable point at which the pendulum is in its obtained hanging position and unstable point in this case refers to a metastable point is satisfied when the pendulum is reversed. Graphically you can get the critical points in the system, we can first obtain which is the equation for calculating the minimum by Figure 20, it can be seen that the minimum occurs when the wave goes from a negative slope to a positive slope, therefore, we are obtained minimum $\pm\pi i$ where $i$ must be odd. Similarly, one can find the maximum, these occur when moving from a positive slope to a negative, therefore the maximum are present in $\pm\pi i$ where $i$ must be zero or even.

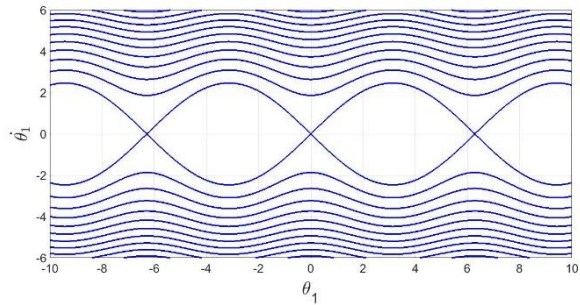
**Figure 21.** Pendulum phase space

Physically each point on the Figure 21 represents a possible system status, explicitly state in which the pendulum would.

As mentioned above the pendulum has two equilibrium points, however, in the phase space endless these points are. This is because there is shown the position like a straight or $R$ space when the real space is a circle that is not topologically equals $R$. Therefore, if space representing geometric positioning system is not topologically equivalent to $R$, for subsequent phase space will not be in $R^2$, since this is the Cartesian product between the geometric space and space representing the speed is generally $R$. Topologically speaking the pendulum has a geometric space $S1$ which is equivalent to a circle and a space velocity $R$, therefore, the phase space $S1 \times R$ equivalent to a cylinder and not to $R^2$. The correct graphic process should be done on a cylinder [18] where its circumference is the geometric representation of the pendulum, this is equivalent to the set of possible positions of the center of mass, while the position along its axis represents the velocity of the center mass, this cylinder is topologically correct space phase. Touring the circumference of the cylinder we can see that there is only one minimum and one maximum diametrically opposed. In Figure 20 you could take energy levels were the result of drawing lines parallel to the axis of the $\theta_1$, straight position, well, now those levels may be taken with circles surrounding the cylinder.

## 8. Conclusions

The formalism of Euler-Lagrange allows a dynamic modeling of the rotary inverted pendulum in a simple way, thanks to that the formalism allows working in a scalar manner. Checking one of its advantages that recommends using the method in case there are rotations in the system.

In this paper, the dynamic modeling of a rotary inverted pendulum has been presented step by step, contributing a complete document to the scientific community that wishes to work with this type of system. In addition, are shown simulations that they validate the equations of motion, comparing the above mentioned with the CAD model realized in SOLIDWORKS, showing valid results. In later studies, these simulations are a good point of reference.

This document differs from the others thanks to the use of the SimMechanics link, which allows multi-body simulations of simple and sophisticated mechanical systems, such as robots, underactuated systems or mechanisms of any kind. It can model sensors, actuators, joints, forces, etc. With the implementation of specialized blocks. Was successfully used this environment in the simulation of the proposed CAD model. Due to its developers it is possible be certain that it is a reliable tool. It is also, very simple to use; this is an advantage to perform simulations of 3D models, since it is not necessary to create platforms for their simulation and it is enough to make a CAD model in any software of design, in this case it was SOLIDWORKS. In addition, MATLAB allows to work with different models in the same environment, in this case, the Simulink-MATLAB environment, where there were realized the simulations of the mathematical model. While the SimMechanics-MATLAB environment was used to simulate the CAD model, it is necessary to clarify that SimMechanics is in the same environment Simulink, which generates comfort at the moment of realizing simulations, Since it is not necessary to change of software for each model.

Additionally, in this article in particular, the analysis of the trajectories of the pendulum is performed, the phase space shown is topologically correct, this being a good contribution to future dynamic, kinematic, chaotic and physical studies of the system.